%% file: main.tex
\documentclass{article}

\usepackage[utf8]{inputenc}
\usepackage{times}
\usepackage{graphicx}
\usepackage{subfigure} \usepackage{natbib}
\usepackage{algorithmic}
\usepackage{hyperref} \usepackage{amsmath}
\usepackage{amssymb}

\newcommand{\DSGF}{\begin{sc}\mbox{DSG-F}\end{sc}}
\newcommand{\DSGS}{\begin{sc}\mbox{DSG-S}\end{sc}}
\newcommand{\HAM}{\begin{sc}SGI\end{sc}}
\newcommand{\KIM}{\begin{sc}SGP\end{sc}}
\usepackage[accepted]{icml2017}

\definecolor{royalblue}{rgb}{0.0, 0.14, 0.4}
\hypersetup{citecolor=royalblue}

\icmltitlerunning{Structured Black Box Variational Inference for Latent Time Series Models}

\begin{document}

\twocolumn[ \icmltitle{Structured Black Box Variational Inference for Latent Time Series Models}

\begin{icmlauthorlist}
\icmlauthor{Robert Bamler}{drp}
\icmlauthor{Stephan Mandt}{drp}
\end{icmlauthorlist}

\icmlaffiliation{drp}{Disney Research, 4720 Forbes Avenue, Pittsburgh, PA 15213,
USA}

\icmlcorrespondingauthor{Robert Bamler}{Robert.Bamler@disneyresearch.com}
\icmlcorrespondingauthor{Stephan Mandt}{Stephan.Mandt@disneyresearch.com}

\icmlkeywords{machine learning, variational inference, time series}

\vskip 0.3in ]

\printAffiliationsAndNotice{}

\begin{abstract}
Continuous latent time series models are prevalent in Bayesian modeling; examples include the Kalman filter, dynamic collaborative filtering, or dynamic topic models. These models often benefit from structured, non mean field variational approximations
that capture correlations between time steps.
Black box variational inference with reparameterization gradients (BBVI) allows us to explore a rich new class of Bayesian non-conjugate latent time series models; however, a naive application of BBVI to such structured variational models would scale quadratically in the number of time steps. We describe a BBVI algorithm analogous to the forward-backward algorithm which instead scales linearly in time. It allows us to efficiently sample from the variational distribution and estimate the gradients of the ELBO. Finally, we show results on the recently proposed dynamic word embedding model, which was trained using our method.
\end{abstract}

\section{Introduction}
\label{sec:motivation}

Continuous latent time series models are popular tools in Bayesian machine learning. One thereby combines multiple copies of a likelihood model through a time series prior, thus enabling the latent variables to drift over time while still sharing statistical strength accross all times~\citep{blei2006dynamic,wang2012continuous,sahoo2012hidden,gultekin_collaborative_2015,charlin2015dynamic,ranganath2015survival,jerfel2016dynamic, bamler2017dynamic}.

Variational Inference (VI) enables  approximate inference for complex models by solving an optimization problem~\citep{jordan_introduction_1999}. One chooses a variational distribution and fits it to the posterior, where the fully-factorized mean field class is the most widely used. However, the standard mean field class is not a good choice when it comes to time series models. Instead, one often uses \emph{structured} variational distributions in time~\citep{wainwright2008graphical,blei2006dynamic}. The perhaps simplest such choice is a multivariate Gaussian with a tridiagonal precision matrix.
This can be seen as using the probabilistic model of the Kalman filter as a variational distribution (variational Kalman filter)~\citep{blei2006dynamic}, and reflects the fact that the prior is a first-order Markov process.

In this paper, we introduce an efficient VI algorithm for fitting such tridiagonal Gaussian variational models to a large class of  complex latent time series models. We draw on Black Box Variational Inference (BBVI) with reparameterization gradients~\citep{salimans2013fixed,kingma2013auto,rezende2014stochastic,ruiz2016generalized}, where one forms a Monte-Carlo estimate of the variational lower bound's gradient. The problem with structured variational distributions is that computing this reparameterization gradient is expensive; a naive implementation involves a dense matrix multiplication and scales as $O(T^2)$, where $T$ is the number of time steps. In this paper, we lay out a general algorithm which gives the reparameterization gradient in $O(T)$. This algorithm can be thought of a variant of the forward-backward algorithm~\citep{murphy2012machine} in the context of BBVI, and relies on a reparameterization procedure that never involves the inversion or multiplication of a dense matrix. We illustrate our approach on  the dynamic word embedding model~\citep{bamler2017dynamic}.

\section{Background and Problem Setting}
We start by specifying our generative and variational models and give an overview of the proposed algorithm.
\label{sec:problem}

\paragraph{Generative model.}
We consider models with time-dependent observations $\mathbf x\equiv(x_1,\ldots, x_T)$, where $T$ is the number of time steps.
At each time step $t\in\{1,\ldots,T\}$, the generative process is described by an arbitrary likelihood function $p(x_t|z_t)$, and $z_t$ is a latent variable.
We furthermore assume that the prior is Markovian.
The joint probability thus factorizes as follows,
\begin{align}
    p(\mathbf x, \mathbf z) = \prod_{t=1}^T p(z_t|z_{t-1})\, p(x_t|z_t),
    \label{eq:model_class}
\end{align}
where $p(z_1|z_0) \equiv p(z_1)$ denotes an arbitrary prior on the first latent variable.
Many models fall into this class. Our goal is an efficient posterior approximation for this model class, using a structured variational approximation and black box inference techniques.

\paragraph{Kalman smoothing revisited.}
Before we describe our approach, we revisit the Kalman smoother \citep{rauch1965maximum} as an efficient algorithm for a particularly simple realization of  Eq.~\ref{eq:model_class} where all conditional distributions are Gaussian.
This is often called a Wiener process (more precisely, it is a Gauss-Markov process).
The prior is Gaussian with tridiagonal precision matrix $\Lambda_0$, and the likelihood is a Gaussian centered around $z_t$ with precision $\tau$.
Thus, the posterior is also Gaussian, $p(\mathbf z|\mathbf x)=\mathcal N(\mathbf z; \boldsymbol\mu, \Lambda^{-1})$, and can be obtained analytically.
One finds $\Lambda = \Lambda_0 + \tau I$ and $\boldsymbol\mu = \tau \Lambda^{-1} \mathbf x$.
Obtaining the posterior modes $\boldsymbol\mu$ involves solving the linear system of equations $\Lambda\boldsymbol\mu = \tau\mathbf x$. For an arbitrary matrix $\Lambda$, this involves $O(T^2)$ operations. However, for the Wiener process, $\Lambda$ is tridiagonal and one can solve for $\boldsymbol\mu$ in linear time using a specialized algorithm.

The forward-backward algorithm~\citep{murphy2012machine} implicitly decomposes $\Lambda=AB$ into a product of two bidiagonal matrices $A$ and $B$, where $A$ is lower triangular and $B$ is upper triangular.
One starts with a forward pass through the data, in which one solves $A\boldsymbol{\tilde\mu} = \tau \mathbf x$ for the auxiliary vector $\boldsymbol{\tilde\mu}$ using forward substitution, followed by a backward pass in which one solves $B\boldsymbol\mu = \boldsymbol{\tilde\mu}$ for $\boldsymbol\mu$ using back substitution. As detailed in section~\ref{sec:backprop}, we use a similar philosophy.

\paragraph{Variational model.}
For a general likelihood model $p(x_t|z_t)$, the exact posterior of Eq.~\ref{eq:model_class} is intractable.
To circumvent this problem, we use BBVI to minimize the KL divergence between a variational distribution $q_{\boldsymbol\lambda}$ and the true posterior.
Here, $\boldsymbol\lambda$ summarizes all variational parameters.
We use a structured variational distribution as our variational model that is motivated by the exact posterior of the analytically solvable model discussed above.
We thus consider a Gaussian with a tridiagonal precision matrix $\Lambda$,
\begin{align}\label{eq:def-q}
    q_{\boldsymbol\lambda}(\mathbf z) \equiv \mathcal N(\mathbf z; \boldsymbol\mu, \Lambda^{-1}).
\end{align}
This can be understood as varying the parameters of a (fictitious) Wiener process so that its posterior approximates the posterior of the true model~\citep{blei2006dynamic}.

As we discuss below, the tridiagonal structure of $\Lambda$ allows us to fit the variational distribution efficiently.
Note that the covariance matrix $\Lambda^{-1}$ is dense, thus encoding long-range correlations between any two time steps.
Modelling correlations is important for many time series applications, in particular when the prior is strong e.g. due to little evidence per time step.

\paragraph{Black box variational inference.}
In BBVI, one fits $q_{\boldsymbol\lambda}(\mathbf z)$ to the posterior $p(\mathbf z|\mathbf x)$, maximizing the evidence lower bound (ELBO) using stochastic gradients. The reparameterization trick amounts to representing the ELBO as
\begin{align} \label{eq:elbo-reparameterized}
    \mathcal L(\boldsymbol\lambda) = \mathbb E_{\boldsymbol\epsilon \sim \mathcal N(0, I)}[\log p(\mathbf x, \mathbf \mathbf f(\boldsymbol\lambda; \boldsymbol\epsilon))] + H[q_{\boldsymbol\lambda}].
\end{align}
The entropy $H[q_{\boldsymbol\lambda}] \equiv -\mathbb E_{q_{\boldsymbol\lambda}}[\log q_{\boldsymbol\lambda}(\mathbf z)]$ is often an analytic function or can be estimated using other tricks.
Here, $\boldsymbol\epsilon$ is a vector of $T$ independent standard-normal distributed random variables, and $\mathbf f\equiv (f_1, \ldots, f_T)$ denotes $T$ functions that are defined such that the random variable $\mathbf f(\boldsymbol\lambda; \boldsymbol\epsilon)$ has probability density $q_{\boldsymbol\lambda}$ (see Section~\ref{sec:forward-prop} below).

In order to implement an efficient BBVI algorithm, one needs to be able to estimate the gradient of $\mathcal L$ efficiently, i.e., in $O(T)$ time.
This involves the following three tasks:
\begin{enumerate}
    \item Efficiently evaluate the entropy $H[q_{\boldsymbol\lambda}]$ (Section~\ref{sec:entropy});
    \item Efficiently evaluate $\mathbf f(\boldsymbol\lambda; \boldsymbol\epsilon)$ (Section~\ref{sec:forward-prop});
    \item Efficiently estimate the reparameterization gradient (Section~\ref{sec:backprop}).
\end{enumerate}

All three of the above tasks can easily be solved efficiently if one chooses a mean field variational distribution, i.e., if $q_{\boldsymbol\lambda}(\mathbf z)$ factorizes over all time steps.
However, the mean field approximation ignores correlations between time steps, which are important in many time series models as discussed above.
In Section~\ref{sec:inference} we address each one of the above tasks individually.

\section{Inference}
\label{sec:inference}
In this section, we give the details of our new black box variational inference algorithm.

\subsection{Evaluating the Entropy}
\label{sec:entropy}

The entropy of a multivariate Gaussian with precision matrix $\Lambda$ is $H[q_{\boldsymbol\lambda}]=-\frac12 \log(\det \Lambda)$, up to an additive constant.
Evaluating the determinant of a general $T\times T$ matrix takes $O(T^3)$ operations in practice.
To avoid this expensive operation, we parameterize the precision matrix via its Cholesky decomposition,
\begin{align} \label{eq:cholesky-lambda}
    \Lambda = B^\top B.
\end{align}
Here, $B$ is an upper triangular $T\times T$ matrix.
Since we restrict $\Lambda$ to have a tridiagonal structure, $B$ is bidiagonal
\citep{kilic_inverse_2013}, i.e., it has the structure
\begin{align}\label{eq:bmatrix}
    B(\boldsymbol\nu, \boldsymbol\omega) = \begin{pmatrix}
        \nu_{1} & \omega_{1} & & & \\
        & \nu_{2} & \omega_{2} & & \\
        & & \ddots & \ddots & \\
        & & & \nu_{T-1} & \omega_{T-1} \\
        & & & & \nu_T
    \end{pmatrix},
\end{align}
with $\nu_t>0\; \forall t\in\{1,\ldots,T\}$.
As the mapping from $B$ to $\Lambda$ is unique, it suffices to optimize the $(2T-1)$ non-zero components of $B$ in order to find the optimal entries of $\Lambda$.
It turns out that we never have to construct the matrix $\Lambda=B^\top B$ explicitly.
The variational parameters ${\boldsymbol\lambda}\equiv(\boldsymbol\mu, \boldsymbol\nu, \boldsymbol\omega)$ are thus the marginal means $\boldsymbol\mu$ (see Eq.~\ref{eq:def-q}) and the non-zero components of $B$.
Using the relation $\det \Lambda = (\det B)^2$, we can evaluate the entropy in linear time,
\begin{align}
    H[q_{\boldsymbol\lambda}]
    = - \sum_{t=1}^T \log\nu_t + const. \label{eq:entropy}
\end{align}

\subsection{Evaluating the reparameterization functions}
\label{sec:forward-prop}

In contrast to the entropy, the expected log-joint in Eq.~\ref{eq:elbo-reparameterized} cannot be evaluated analytically for a general model.
We obtain an unbiased gradient estimator $\mathbf{\hat g}$ of the expected log-joint by drawing $S$ independent samples $\boldsymbol\epsilon^{[s]} \sim \mathcal N(0,I)$ for $s\in\{1,\ldots, S\}$.

For what follows, let $\lambda_i$ denote any of the $(3T-1)$ variational parameters.
Using the chain rule, the estimate of the gradient of the expected log-joint with respect to $\lambda_i$ is
\begin{align}
    \hat g_i &\equiv \frac{\partial}{\partial \lambda_i}\left[ \frac{1}{S}\sum_{s=1}^S \log p(\mathbf x, \mathbf f(\boldsymbol\lambda, \boldsymbol\epsilon^{[s]}))\right] \nonumber\\
    &= \frac{1}{S}\sum_{s=1}^S \sum_{t=1}^T  \gamma_t^{[s]} \frac{\partial f_{t}({\boldsymbol\lambda}; \boldsymbol\epsilon^{[s]})}{\partial \lambda_i} \label{eq:reparam-gradient}
\end{align}
where we defined
\begin{align}
    \gamma_t^{[s]} &\equiv \left. \frac{\partial\log p(\mathbf x,\mathbf z^{[s]})}{\partial z_t^{[s]}}\right|_{\mathbf z^{[s]} = \mathbf f(\boldsymbol\lambda, \boldsymbol\epsilon^{[s]})} \label{eq:def-gamma}\\
    &= \frac{\partial [ \log p(z_t^{[s]}|z_{t-1}^{[s]}) \!+\! \log p(z_{t+1}^{[s]}|z_t^{[s]}) \!+\! \log p(x_t|z_t^{[s]})]}{\partial z_t^{[s]}}.\nonumber
\end{align}
To further simplify Eq.~\ref{eq:reparam-gradient} we  specialize the reparameterization function to our variational model. Using Eqs.~\ref{eq:def-q} and \ref{eq:cholesky-lambda}, we find
\begin{align}\label{eq:def-f}
    \mathbf f(\boldsymbol\mu,\boldsymbol\nu, \boldsymbol\omega; \boldsymbol\epsilon^{[s]}) = \boldsymbol\mu + B(\boldsymbol\nu,\boldsymbol\omega)^{-1} \boldsymbol\epsilon^{[s]}.
\end{align}
Here, $B^{-1}$ is a dense (upper triangular) $T\times T$ matrix.
Instead of computing the inverse, we evaluate the term $\mathbf y^{[s]}\equiv B^{-1}\boldsymbol\epsilon^{[s]}$ on the right-hand side of Eq.~\ref{eq:def-f} by solving the linear system $B\mathbf y^{[s]}=\boldsymbol\epsilon^{[s]}$ for $\mathbf y^{[s]}$ using back substitution.
This takes only $O(T)$ operations due to the bidiagonal structure of $B$, see Eq.~\ref{eq:bmatrix}.
We can therefore evaluate Eq.~\ref{eq:def-gamma} in $O(T)$ time for each sample $s\in\{1,\cdots,S\}$.

\subsection{Estimating the Reparameterization Gradient}
\label{sec:backprop}
Last, we show how to efficiently estimate the reparameterization gradient in $O(T)$ time.
In this subsection we describe an efficient method to evaluate the Jacobian $\partial\mathbf f/\partial \boldsymbol\lambda$ that appears on the right-hand side of Eq.~\ref{eq:reparam-gradient}.

A naive evaluation of all gradient estimates $\hat g_i$ would require $O(S\times T^2)$ operations: the sums on the right-hand side of Eq.~\ref{eq:reparam-gradient} run over $S\times T$ terms, and the number of variational parameters $\lambda_i$ for which the gradient estimate has to be evaluated grows at least linearly in $T$ for a reasonably flexible variational distribution.

We use the following trick to  evaluate all gradient estimates in linear time in $T$.
For any invertible matrix $B$ that depends on some parameter $\lambda_i$, we have
\begin{align}
    0 = \frac{\partial I}{\partial \lambda_i}
    = \frac{\partial (B B^{-1})}{\partial \lambda_i}
    = \frac{\partial B}{\partial \lambda_i} B^{-1} + B \frac{\partial B^{-1}}{\partial \lambda_i}.
\end{align}
Solving for $\partial B^{-1}/\partial \lambda_i$, we obtain
\begin{align}\label{eq:derivative-inverse}
    \frac{\partial B^{-1}}{\partial \lambda_i} = - B^{-1} \frac{\partial B}{\partial \lambda_i} B^{-1}.
\end{align}
Eq.~\ref{eq:derivative-inverse} expresses the derivative of the dense (upper triangular) matrix $B^{-1}$ in terms of the derivative of the bidiagonal matrix $B$.
In fact, both $\partial B/\partial \nu_t$ and $\partial B/\partial \omega_t$ are sparse matrices with only a single non-zero entry, see Eq.~\ref{eq:bmatrix}.
The dense matrix $B^{-1}$ still appears on the right-hand side of Eq.~\ref{eq:derivative-inverse} but we avoid evaluating it explicitly.
Instead, we again solve a bidiagonal linear system of equations.
Combining Eqs.~\ref{eq:reparam-gradient}, \ref{eq:def-f} and \ref{eq:derivative-inverse}, we obtain the formal expressions
\begin{align}
    \hat g_{\mu_t} &= \frac{1}{S} \sum_{s=1}^S \gamma_t^{[s]}; \label{eq:grad-elbo-mu} \\
    \hat g_{\nu_t} &= - \frac{1}{S} \sum_{s=1}^S \underbrace{((\boldsymbol\gamma^{[s]})^\top B^{-1})_t}_{y^{\prime[s]}_t} \, \underbrace{(B^{-1} \boldsymbol\epsilon^{[s]})_t}_{y^{[s]}_t}; \label{eq:grad-elbo-nu} \\
    \hat g_{\omega_t} &= - \frac{1}{S} \sum_{s=1}^S \underbrace{((\boldsymbol\gamma^{[s]})^\top B^{-1})_t}_{y^{\prime[s]}_t}\, \underbrace{(B^{-1} \boldsymbol\epsilon^{[s]})_{t+1}}_{y^{[s]}_{t+1}} \label{eq:grad-elbo-omega}
\end{align}
where bold face $\boldsymbol\gamma^{[s]} \equiv (\gamma^{[s]}_1,\ldots, \gamma^{[s]}_T)^\top$ denotes a column vector of $T$ derivatives of the log-joint as defined in Eq.~\ref{eq:def-gamma}.
Instead of computing the inverse $B^{-1}$ in Eqs.~\ref{eq:grad-elbo-nu}--\ref{eq:grad-elbo-omega}, we obtain the vectors $(\mathbf y^{\prime[s]})^\top \equiv (\boldsymbol\gamma^{[s]})^\top B^{-1}$ and $\mathbf y^{[s]} \equiv B^{-1} \boldsymbol\epsilon^{[s]}$ by solving the linear systems $B^\top \mathbf y^{\prime[s]} = \boldsymbol\gamma^{[s]}$ and $B \mathbf y^{[s]} = \boldsymbol\epsilon^{[s]}$, respectively, in $O(T)$ time using the bidiagonal structure of $B$.
Since $B^\top$ is lower-diagonal and $B$ is upper-diagonal these operations carry-out a forward and a backward pass through the time steps, respectively.
This is similar to the forward-backward algorithm for the Wiener process with Gaussian likelihood discussed in Section~\ref{sec:problem}.

Eqs.~\ref{eq:grad-elbo-mu}--\ref{eq:grad-elbo-omega} conclude the derivation of the gradient estimator for the expected log-joint.
The gradient of the ELBO with respect to $\nu_t$ contains an additional term $-1/\nu_t$ due to the entropy, see Eq.~\ref{eq:entropy}.

\input{experiments}

\bibliography{references} \bibliographystyle{icml2017}

\end{document}

%% file: experiments.tex
\begin{figure*}[tb!]
\begin{center}
\centerline{\includegraphics[width=\textwidth]{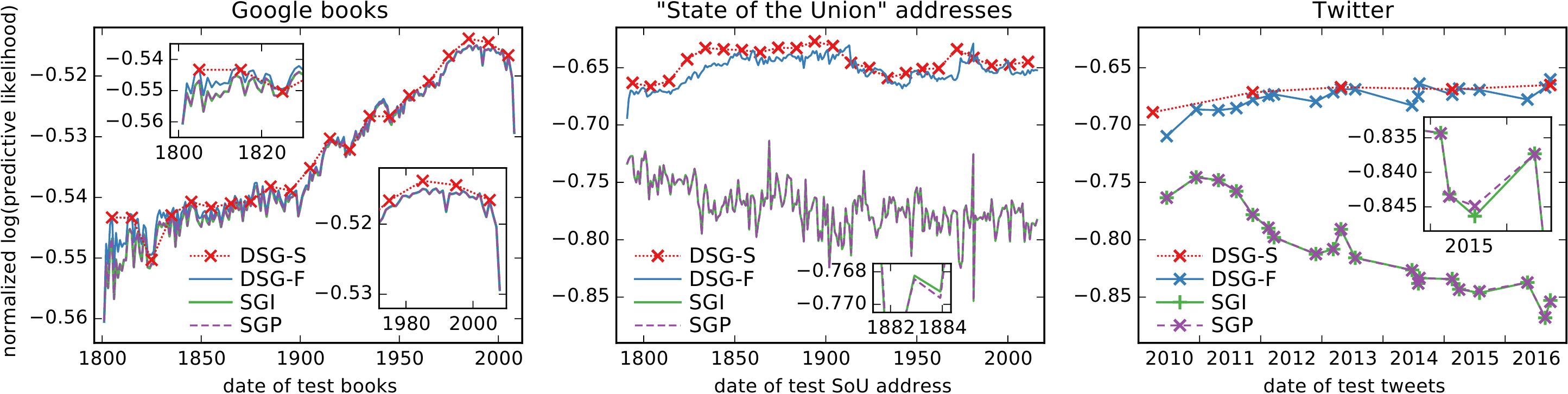}}
\caption{Predictive log-likelihoods for two proposed versions of the dynamic skip-gram model (DSG-F \& DSG-S) and two competing methods \HAM{}~\citep{hamilton2016diachronic} and \KIM{}~\citep{kim2014temporal} on three different corpora (high values are better). The proposed structured BBVI algorithm (red crosses) outperforms all baselines.
}
\label{fig:pred-log-likelihoods}
\end{center}
\vskip -0.2in
\end{figure*}

\section{Experiments}
\label{sec:experiments}

To demonstrate the benefits of our approach, we summarize selective experimental results from work on dynamic word embeddings, which used our proposed BBVI algorithm.

\paragraph{Dynamic Skip-Gram Model.}
\citet{bamler2017dynamic} proposed the dynamic skip-gram model as a generalization of the skip-gram model to time-stamped text corpora~\citep{mikolov_distributed_2013}. The model, thus, represents words as latent trajectories in an embedding space, and is able to track how words change their meaning over time. 

To train the model, one first summarizes the text corpus in terms of certain sufficient statistics, using a finite vocabulary. For each time stamp, one builds co-occurrence matrices of words and their surrounding words. In addition, the model requires so-called \emph{nagative} examples, expressing the likelihood of co-occurrance by mere chance. The model then models these statistics in terms of the geometrical arrangement of certain latent word and context embedding vectors. The prior over these embedding vectors is a latent Ornstein-Uhlenbeck process (a Wiener process in the presence of a restoring force), enforcing a continuous drift over time. For more details, we refer to~\citep{bamler2017dynamic}.


\paragraph{Algorithms and Baselines.} 
{\bf \HAM{}} denotes the non-Bayesian skip-gram model with independent random initializations of word vectors~\citep{mikolov_distributed_2013,hamilton2016diachronic}. 
{\bf \KIM{}} denotes the same approach as above, but with word and context vectors being pre-initialized with the values from the previous year, as in~\citep{kim2014temporal}. Both approaches have no dynamical priors and hence no overhead and just scale linearly with $T$.
{\bf \DSGF{}} denotes the dynamic skip-gram filtering algorithm, proposed in~\citep{bamler2017dynamic}, which also runs in $O(T)$, but uses the mean field approximation and only uses information form the past. Finally, {\bf \DSGS{}} denotes the dynamic skip-gram smoothing algorithm. This is the $O(T)$ algorithm proposed in this paper, applied to the dynamic skip-gram model.

\paragraph{Data.}
 We used data from the  Google books corpus~\citep{michel_quantitative_2011} from the last two centuries ($T=209$). We also used the ``State~of~the~Union'' ({SoU}) addresses of U.S.~presidents, which spans more than two centuries. Finally, we used a {Twitter} corpus of news tweets for $21$ randomly drawn dates from $2010$ to $2016$. Details on hyperparameters and preprocessing are given in~\citep{bamler2017dynamic}.

\paragraph{Results.}
We focus on the quantitative results of~\citep{bamler2017dynamic}, showing show that the dynamic skip-gram smoothing algorithm (described and generalized in this paper) generalizes better to unseen data than all baselines at similar run time. We thereby analyze held-out predictive likelihoods on word-context pairs at a given time $t$, where $t$ is excluded from the training set. 

The predictive likelihoods as a function of time are shown in Figure~\ref{fig:pred-log-likelihoods}.
On all three corpora,
differences between the two implementations of the static model (\HAM{} and \KIM{}) are small, which suggests that pre-initializing the embeddings with the previous result seems to have little impact on the predictive power.
Log-likelihoods for the skip-gram filter (DSG-F) grow over the first few time steps as the filter sees more data, and then saturate. 
Skip-gram smoothing (this paper) further outperforms skip-gram filtering. 

To conclude, we stress that a structured BBVI algorithm with quadratic instead of linear run time in $T$ would be impractical. We therefore hope that our structured reparameterization trick will spur further research on complex latent time series models.